\documentclass[11pt,a4paper]{article}
\usepackage[table]{xcolor}
\usepackage[hyperref]{acl2021}
\usepackage{tabularx}

\usepackage{times}
\usepackage{latexsym}
\usepackage{diagbox}

\usepackage{array}

\usepackage{microtype}
\usepackage{amsmath}
 \usepackage{multirow}
\usepackage[position=bottom]{subfig}
\usepackage{booktabs}
\usepackage{CJKutf8}
\newcommand{\minus}{\scalebox{0.75}[1.0]{$-$}}
\def\aclpaperid{473}
\usepackage{xspace}

\usepackage{pgf}
\usepackage{collcell}
\usepackage{booktabs}
\usepackage{tcolorbox}

\newcommand*{\MinNumber}{-0.5}%
\newcommand*{\MaxNumber}{0.5}%

\newcommand{\ApplyGradient}[1]{%
  \pgfmathsetmacro{\PercentColor}{100.0*((#1+\MaxNumber)/(\MaxNumber-\MinNumber))}%
  \edef\x{\noexpand\cellcolor{red!\PercentColor}}\x\textcolor{white}{#1}%
}
\newcolumntype{R}{>{\collectcell\ApplyGradient}{r}<{\endcollectcell}}
\aclfinalcopy

\newcommand{\eg}{\textit{e.g.,}\xspace}
\newcommand{\ie}{\textit{i.e.,}\xspace}
\newcommand{\Fone}{F\textsubscript{1}\xspace}

\title{Cross-Lingual Transfer in Zero-Shot Cross-Language Entity Linking}

\author{Elliot Schumacher \hspace{1em}
        James Mayfield \hspace{1em}
        Mark Dredze \hspace{1em} \\ 
        Johns Hopkins University \\
        \texttt{eschumac@cs.jhu.edu} \hspace{1em} \texttt{mayfield@jhu.edu} \hspace{1em} \texttt{mdredze@cs.jhu.edu}}

\date{}
\begin{document}

\maketitle

\begin{abstract}
Cross-language entity linking grounds mentions written in several languages to a monolingual knowledge base.
We use a simple neural ranking architecture for this task that uses multilingual BERT representations of both the mention and the context as input, so as to explore the ability of a transformer model to perform well on this task.
We find that the multilingual ability of BERT leads to good performance in monolingual and multilingual settings.
Furthermore, we explore zero-shot language transfer and find surprisingly robust performance. 
We conduct several analyses to identify the sources of performance degradation in the zero-shot setting.  Results indicate that while multilingual transformer models transfer well between languages, issues remain in disambiguating similar entities unseen in training.
\end{abstract}

\section{Introduction}

Entity linking grounds named entities mentioned in text, such as \textit{Chancellor}, to a reference knowledge base (KB) or ontology entry, such as \textit{Angela Merkel}.   Historically, entity linking work focused on English documents and knowledge bases, but subsequent work expanded the task to consider multiple languages~\cite{McNamee2011}. In cross-language entity linking, entities in a set of multilingual documents is linked to a KB in a single language.
The TAC KBP shared task~\cite{Ji2015}, for example, links mentions in Chinese and Spanish documents with an English KB.
Success in building cross-language linking systems can be helpful in tasks such as discovering all documents relevant to an entity, regardless of language.

\begin{figure*}[!htb]
    \centering
    \begin{minipage}{.28\textwidth}
        \centering
    \begin{tcolorbox}[width=4.2cm,
                  boxsep=0pt,
                  arc=0pt,
                  colback=white,
                  left=2pt,
                  right=2pt,
                  top=2pt,
                  bottom=2pt
                  ]
                  ...  el jefe de la \textbf{Oficina de la Presidencia} \textit{(m.01p1k, ORG)},
                Aurelio Nuño y ...
    \end{tcolorbox}
                \centering
\begin{tabular}{@{}r|p{2.8cm}@{}}
name & \textit{President of Mexico} (m.01p1k) \\ \midrule
desc. & \textit{The President of the United  ...} \\ \midrule
type & government\_office
\end{tabular} 
    \end{minipage}%
    \begin{minipage}{0.72\textwidth}
        \centering
        \includegraphics[width=0.98\textwidth]{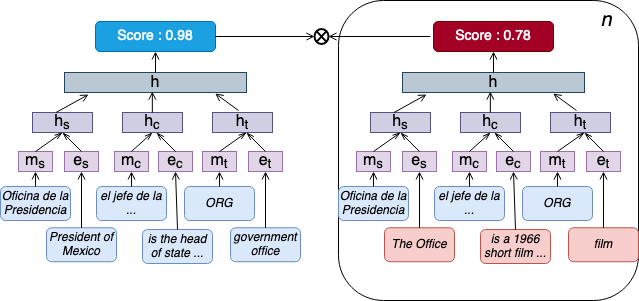}

    \end{minipage}
\caption{Example Spanish mention \textit{Oficina de la Presidencia}, which is a link to entity \textit{President of Mexico}, and the architecture for our neural ranker, using that example and a negatively-sampled entity \textit{The Office}.}
\label{fig:ex_arch}
\end{figure*}

Successfully linking a mention across languages requires adapting several common entity linking components to the cross-language setting.
Consider the example in Figure \ref{fig:ex_arch}, which contains the
Spanish mention \textit{Oficina de la Presidencia}, a reference to the entity \textit{President of Mexico} in an English KB. To link the mention to the relevant entity we must compare the mention text and its surrounding textual context in Spanish to the English entity name and entity description, as well as compare the mention and entity type.  Previous work has focused on transliteration or translation approaches for name and context~\cite{McNamee2011, pan2015unsupervised}, or leveraging large amounts of cross-language information~\cite{Tsai2016} and multilingual embeddings~\cite{upadhyay-etal-2018-joint}.

Since this work emerged, there have been major advances in multilingual NLP~\cite{wu-dredze-2019-beto, pires-etal-2019-multilingual}. Mainstream approaches to multilingual learning now use multilingual encoders, trained on raw text from multiple languages~\cite{devlin-etal-2019-bert}. These models, such as multilingual BERT or XMLR~\cite{conneau2019unsupervised}, have achieved impressive results on a range of multilingual NLP tasks, including part of speech tagging~\cite{tsai-etal-2019-small}, parsing~\cite{wang-etal-2019-cross, kondratyuk-straka-2019-75}, and semantic similarity~\cite{lo-simard-2019-fully,reimers-gurevych-2019-sentence}.

We propose to leverage text representations with 
multilingual BERT~\cite{devlin-etal-2019-bert} for cross-language entity linking to handle
the mention text, entity name, mention context and entity description\footnote{Our code is available at \url{https://github.com/elliotschu/crosslingual-el}}. We use a neural ranking objective and a deep learning model to combine these representations, along with a 
one-hot embedding for the entity and mention type,
to produce a cross-language linker.  We use this ranking architecture to highlight the ability of mBERT to perform on this task without a more complex architecture.
Although previous work tends to use multilingual encoders for one language at a time, \eg train a Spanish NER system with mBERT, we ask: can our model effectively link entities {\em across} languages? 
We find that, somewhat surprisingly, 
our approach does exceedingly well;
scores are comparable to previously reported best results that are trained on data not available to our model (they have access to non-English names).
Next, we consider a multilingual setting, in which a single system is simultaneously trained to link mentions in multiple languages to an English KB. 
Previous work~\cite{upadhyay-etal-2018-joint} has shown that multilingual models can perform robustly on cross-language entity linking. Again, we find that, surprisingly, a model trained on multiple languages at once does about as well, or in some cases better, than the same model trained separately on every language.

These encouraging results lead us to explore the challenging task of zero-shot training, in which we train a model to link single language documents  (\eg English) to an English KB, but apply it to unseen languages (\eg Chinese) documents. 
While the resulting model certainly does worse on a language that is unobserved, the reduction in performance is remarkably small.
This result leads us to ask: 1) Why do zero-shot entity linking models do so well? 2) What information is needed to allow zero-shot models to perform as well as multilingually trained models? 
Using a series of ablation experiments we find that correctly comparing the mention text and entity name is the most important component of an entity linking model. Therefore, we propose an auxiliary pre-training objective to improve zero-shot performance. However, we find that this text-focused approach does not improve performance significantly.
Rather, we find that much of the remaining loss comes not from the language transfer, but from mismatches of entities mentioned across the datasets. This suggests that future work on the remaining challenges in zero-shot entity linking should focus on topic adaptation, instead of improvements in cross-lingual representations.

In summary, this paper uses a simple ranker to explore effective cross-language entity linking with multiple languages. We demonstrate its effectiveness at zero-shot linking, evaluate a pre-training objective to improve zero-shot transfer, and lay out guidelines to inform future research on zero-shot linking.

\section{Cross-Language Entity Linking}
A long line of work on entity linking has developed standard models to link textual mentions to entities in a KB~\cite{dredze2010entity,durrett-klein-2014-joint,gupta-etal-2017-entity}. The models in this area have served as the basis for developing multilingual and cross-language entity linking systems, and they inform our own model development.  We define {\bf multilingual} to mean a model that can operate on mentions from more than one language at the same time (link both English and Chinese mentions to an ontology) and {\bf cross-language} to refer to linking mentions in one language (\eg Spanish) to an ontology in another (\eg English).

A common approach to cross-language entity linking is to use transliteration data to transform non-English mentions into English strings.
Early transliteration work~\cite{McNamee2011} uses a transliteration corpus to train a support vector machine ranker, which uses common entity linking features such as name and context matching, co-occurring entities, and an indicator for NIL (no matching candidate.)  \citet{pan2017cross} uses transliteration data for a set of 282 languages to generate all possible combinations of mentions. 
A related approach is to use machine translation to translate a document into English, and then use an English entity linker.  However, an MT system may not be available, and it further needs a specialized name module to properly translate entity names. Several systems from the TAC 2015 KBP Entity Discovery and Linking task~\cite{Ji2015} translate non-English documents into English, then use standard Entity Linking systems.

Cross-language Wikification is a closely related task, which uses links within Wikipedia, combined with equivalent pages in other languages to train an entity linker with Wikipedia as the KB. This approach typically uses English Wikipedia as the KB, though it could use a KB in other languages. ~\citet{Tsai2016} use a two-step linking approach, first using an IR-based triage system (which we also use).  Second, they use a candidate ranking step based on a linear ranking SVM model with several features, including contextual, document, and coreference.

The most closely related work to our own is that of ~\citet{upadhyay-etal-2018-joint}, who use multilingual embeddings as the basis for their representations, and Wikipedia as training data.  They use FastText~\cite{bojanowski2017enriching,smith2017offline} to align embeddings across languages, and a small dictionary to identify alignments.  They pass these representations through a convolutional neural network to create a mention representation. They in turn use the other mention representations in the document to create a contextual representation, and also use a separate type vector. They train their network on hyperlinks from multiple languages in Wikipedia. Before the ranking step, they use a triage system similar to that of \citet{Tsai2016}. They evaluate on several entity linking datasets, including \textbf{TAC}.  As their system only uses English Wikipedia as the KB, they set all mentions that link to a entity outside of Wikipedia to NIL; this results in a different evaluation setup than we need for our work. Their results show that training on all languages, instead of monolingual or bilingual training, generally performs best.  For zero-shot entity linking, they train on English language Wikipedia.  They find that their performance is heavily dependent on a prior probability derived from the triage system -- otherwise, there is a large drop in performance.

\citet{rijhwani2019zero} investigate zero-shot entity linking on low-resource languages. They propose a model consisting of a similarity model using encoders separately trained on high-resource language mentions, related to the low-resource language, and English entities. They then use the high-resource language as a pivot language for low resource language mentions, allowing them to score mentions in an unseen language.
\citet{raiman2018deeptype} consider multilingual entity linking, in which they use a KB in the same language as the mention, but exploit multilingual transfer for the model's type system. They formulate a type system as a mixed integer problem, which they use to learn a type system from knowledge graph relations. 

\begin{table}[htb!]
    \centering
\begin{tabular}{@{}cl|cccc@{}} \toprule
\multirow{6}{*}{\rotatebox[origin=c]{90}{en}} 
& NN & 0.195 & 0.463 & 0.550 & 0.502 \\
& Mono & 0.586 & \textbf{0.703} & 0.619 & \textbf{0.658} \\
& MultiDS & 0.509 & 0.873 & 0.478 & 0.618 \\
& Multi & \textbf{0.602} & 0.691 & \textbf{0.626} & 0.655 \\
& \textit{MultiOr} & \textit{0.654} & \textit{0.773} & \textit{0.641} & \textit{0.703} \\
& \textit{Tri} & --- & \textit{0.736} & \textit{0.738} & \textit{0.737} \\\midrule
\multirow{6}{*}{ \rotatebox[origin=c]{90}{zh} } &  NN & 0.207 & 0.889 & 0.449 & 0.597 \\
& Mono & 0.709 & \textbf{0.867} & 0.728 & 0.791 \\
& MultiDS & \textbf{0.733} & \textbf{0.867} & \textbf{0.746} & \textbf{0.801} \\
& Multi & 0.730 & 0.862 & 0.735 & 0.793 \\
& \textit{MultiOr} & \textit{0.828} & \textit{0.950} & \textit{0.812} & \textit{0.876} \\ 
& \textit{Tri} & --- & \textit{0.854} & \textit{0.809} & \textit{0.831} \\\midrule
\multirow{6}{*}{ \rotatebox[origin=c]{90}{es}  } & NN & 0.214 & 0.508 & 0.552 & 0.529 \\
& Mono & 0.595 & \textbf{0.921} & 0.587 & 0.714 \\
& MultiDS & 0.604 & 0.918 & 0.590 & 0.718 \\
& Multi & \textbf{0.652} & 0.918 & \textbf{0.625} & \textbf{0.744} \\
& \textit{MultiOr} & \textit{0.691} & \textit{0.936} & \textit{0.655} & \textit{0.770} \\ 
& \textit{Tri} & --- & \textit{0.804} & \textit{0.804} & \textit{0.804} \\\midrule \midrule
\multicolumn{2}{c|}{ \textbf{Model} } & \textbf{micro} & \textbf{prec}. & \textbf{recall} & \textbf{\Fone} \\  \midrule \midrule
\multirow{3}{*}{ \rotatebox[origin=c]{90}{ar}  } & NN & 0.171 & 0.414 & 0.602 & 0.491 \\
& Mono & \textbf{0.660} & \textbf{0.683} & \textbf{0.816} & \textbf{0.743} \\
& Multi & 0.637 & 0.661 & 0.778 & 0.715 \\ \midrule
\multirow{3}{*}{ \rotatebox[origin=c]{90}{fa}  } & NN & 0.330 & 0.694 & 0.734 & 0.714 \\
& Mono & 0.702 & 0.780 & 0.881 & 0.827 \\
& Multi & \textbf{0.762} & \textbf{0.817} & \textbf{0.919} & \textbf{0.863} \\ \midrule
\multirow{3}{*}{ \rotatebox[origin=c]{90}{ko}  } & NN & 0.269 & 0.816 & 0.597 & 0.690 \\
& Mono & 0.752 & 0.832 & 0.861 & 0.846 \\
& Multi & \textbf{0.805} & \textbf{0.850} & \textbf{0.902} & \textbf{0.875} \\ \midrule
\multirow{3}{*}{ \rotatebox[origin=c]{90}{ru}  } & NN & 0.358 & 0.841 & 0.529 & 0.649 \\
& Mono & 0.694 & 0.834 & 0.843 & 0.837 \\
& Multi & \textbf{0.740} &\textbf{0.865} &\textbf{0.876} & \textbf{0.871} \\ \bottomrule
\end{tabular} 
    \caption{Micro-avg. precision, precision, recall, and \Fone for \textbf{TAC} and \textbf{Wiki} datasets. In a majority of languages, the \textbf{Multi} model outperforms the \textbf{Mono} model.}
    \label{tab:mono_multi}
\end{table}
\section{Entity Linking Model} \label{sec:model}

We propose a cross-language entity linker based on a pointwise neural ranker that scores a mention $m$ and entity $e$ pair, adapting from an architecture discussed in~\citet{dehghani2017neural}. Unlike a classification architecture, a ranking architecture is able to score previously unseen entities. As is standard, we use a two stage system: triage followed by ranking; this reduces the number of entities that must be ranked, and results in better performance. Our system is shown in Figure \ref{fig:ex_arch}.  We select this architecture so as to focus on the ability of multilingual transformers to handle this task.

The ranker takes as input information about the mention and entity: 1) the mention string and entity name; 2) the context of the mention and entity description; and 3) the types of the mention and entity. 
We represent the mention string, entity name, mention context and entity description using a pre-trained multilingual deep transformer encoder~\cite{devlin-etal-2019-bert}, while the mention and entity type are represented as one-hot embeddings. 
We describe the multilingual representation, model architecture and training procedure.

\subsection{Multilingual Representations}\label{sec:rep}



We use multilingual BERT (mBERT)~\cite{devlin-etal-2019-bert}\footnote{We found that XLM-R \cite{conneau2019unsupervised} performed similarly and only report results on mBERT.}, which has been shown to create effective multilingual representations for downstream NLP tasks \cite{wu-dredze-2019-beto}.
Consider the Spanish example in Figure~\ref{fig:ex_arch}. First, we create a representation of the mention text $m_s$, \textit{Oficina de la Presidencia}, by creating an mBERT representation of the entire sentence, selecting the lowest layer representations of each of the mention's sub-words,\footnote{We experimented with several BERT layers and found this to be the best performing on the \textbf{TAC} development set.} and form a single representation using max pooling. We create a representation of the entity name $e_s$, \textit{President of Mexico} in the same way, although there is no surrounding context as in a sentence.

For the mention context $m_c$ we select the surrounding sentences up to BERT's $512$ sub-word limit, positioning the mention in the middle, and pass the text to BERT, using the resulting top layer of the {\tt [CLS]} token.
We create a similar representation for the entity context $e_c$ from the definition or other text in the KB, using the first 512 subword tokens from that description. For the mention type $m_t$ and entity type $e_t$ we create one-hot embeddings, omitting ones that do not occur more than 100 times in the training set.

\subsection{Architecture}
We feed the representations of the name ($m_s$ and $e_s$), context ($m_c$, $e_c$) and type ($m_t$, $e_t$) into a neural ranker. Each of these three pairs is passed into distinct multilayer perceptrons (MLPs), which each produce an embedding that captures the similarity between each type of information.
For example, we input $m_s$ and $e_s$ into a text-specific hidden layer, which produces a combined representation $r_s$.  The same is done for the context and type representations, producing representations $r_c$ and $r_t$, respectively.  These three representations are then fed into a final MLP, which produces a final score ($[-1,1]$.) The entire network is jointly trained with the ADAM optimizer and a ranking objective.  We apply dropout at every layer, use ReLu as the intermediate activation function, and Tanh for the final layer.  While additional features such as entity salience are likely useful for this task, we chose to restrict our model as much as possible to use only text features.  This focuses on mBERT's multilingual ability, and allows for easier adaptation to new KBs than with KB-specific features.

\subsection{Model Training}
\label{sec:training}
We learn the parameters $\theta$ of our scoring function $S$ using a pairwise approach; this allows us to train our model without annotated scores. Our ranker scores a mention $m$ and positive entity $e_+$ pair, and separately scores the same mention paired with $n$ sampled negative entities $e_-$. We apply the hinge loss between our correct entity and the highest scoring negative entity,
\begin{multline*}
    \mathit{L}(\theta) = \mathbf{max}\{0,\epsilon-(S(\{m,e_{+}\};\theta)-\\ \mathbf{max}\{S(\{m,e_{0-}\};\theta) \dots 
            S(\{m,c_{n-}\};\theta)\}\}
\end{multline*}
We jointly train all components of the network, including the positive and negative portions of the network. The major benefit of this pairwise approach is that it does not rely on annotated scores, but instead uses negative sampling to train the ranker. We tested random combinations of hidden layer sizes and dropout rates to find the best configuration (see Appendix A for parameter selection details). 

\section{Datasets}
We conduct our evaluation on two cross-language entity linking datasets.  We predict NILs by applying a threshold; mentions where all entities are below a given threshold are marked as NIL. We evaluate all models using the evaluation script provided by~\citet{Ji2015}, which reports Precision, Recall, \Fone, and Micro-averaged precision.
For implementation details, please see the appendix. 

{\bf TAC.}
The 2015 TAC KBP Entity Linking dataset~\citep{Ji2015} consists of newswire and discussion form posts in English, Spanish, and Mandarin Chinese linked to an English KB.  We use their evaluation set, and provide a comparison to the numbers noted in~\citet{Ji2015}.  The referenced systems had access to non-English language KB text which we exclude, and thus are a goal rather than a baseline.  Later papers, such as~\citet{upadhyay-etal-2018-joint}, also use this dataset but only for evaluation, instead training on Wikipedia and treating mentions that are linked to TAC entities without Wikipedia links as NIL. Therefore, we cannot compare our evaluation to this work.

{\bf Wiki.}
We created a cross-language entity linking dataset from Wikipedia links~\cite{pan2017cross} that includes Korean, Farsi, Arabic, and Russian. A preprocessed version of Wikipedia has links in non-English Wikipedia pages to other non-English pages annotated with that link and an English page link if a corresponding page was available. From these annotations we created a dataset consisting of non-English mentions linked to English-language entities (Wikipedia page) using English Wikipedia as the KB. We consider this to be silver-standard data because--unlike the \textbf{TAC} dataset--the annotations have not been reviewed by annotators. Since we do not have a separate development set for this dataset, we apply the hyperparameters selected on \textbf{TAC} development data to this dataset.

\textbf{Triage.}
We assume gold-standard mention boundaries in our analysis. We use the triage system of~\citet{upadhyay-etal-2018-joint}, which is largely based on work in~\citet{Tsai2016}. This allows us to score a smaller set of entities for each mention as opposed to the entire KB. For a give mention $m$, a triage system will provide a set of $k$ candidate entities ${e_1  \dots e_k}$.   The system uses Wikipedia cross-links to generate a prior probability $\mathtt{P}_\mathtt{prior}(e_i|m)$ by estimating counts from those mentions.  This prior is used to provide the top $k$ English Wikipedia page titles for each mention ($k=10$ for \textbf{TAC} and $k=100$ for \textbf{Wiki}).

\section{Model Evaluation} \label{sec:model_eval}
We consider several different training and evaluation settings to explore the multilingual ability of transformers on this task.
Recent studies suggest that multilingual models can achieve similar or even better performance on cross-language entity linking~\cite{upadhyay-etal-2018-joint}.  Other work~\cite{mueller2020sources} has shown that this is not always the case. Therefore, we begin by asking: does our linker do better when trained on all languages (multilingual cross-language) or trained separately on each individual language (monolingual cross-language)?

We train our model on each of the 7 individual languages in the two datasets (noted as \textbf{Mono}).  Next, we train a single model for each dataset (3 languages in \textbf{TAC}, 4 in \textbf{Wiki}, each noted as \textbf{Multi}). \textbf{Mono} and \textbf{Multi} share the exact same architecture - there are no multilingual adjustments made, and the model contains no language-specific features. As \textbf{Multi} uses data available in all languages and thus has more training data than \textbf{Mono}, we include a model that is trained on a randomly-sampled subset of the multilingual training data that set to match the training size of \textbf{Mono} (\textbf{MultiDS}) . For \textbf{TAC} \textbf{Multi} models, we also report results using a candidate oracle instead of triage (\textbf{Multi+Or}), where the correct entity is always added to the candidate list.  For all \textbf{Mono} and \textbf{Multi}-based models we report the average of three runs.  The metric-specific standard deviations were all small, with all but one at or below 0.017.  We note the best performing architecture from \cite{Ji2015} as \textbf{Tri}, again noting that those systems have access to non-English text.  We also evaluate a simple nearest neighbor model (noted as \textbf{NN}).  This model scores each mention-entity pair using the cosine similarity between the mention name representation $m_s$ and the entity representation $e_s$, and selects the highest-scoring pair.


Table \ref{tab:mono_multi} shows that for \textbf{TAC} there is a small difference between the \textbf{Mono} and \textbf{Multi} models. 
For \textbf{Wiki} the difference is often larger. \textbf{Multi} often does better than \textbf{Mono}, suggesting that additional training data is helpful specifically for languages (\eg Farsi) with smaller amounts of data. Overall, these results are encouraging as they suggest that a single trained model for our system can be used for cross-language linking for multiple languages. This can reduce the complexity associated with developing, deploying and maintaining multiple models in a multilingual environment.  For some models, the \textbf{Multi} improvement may be due to additional data available, as shown in the difference in performance between \textbf{Multi} and \textbf{MultiDS} (\eg Spanish \Fone \textbf{Multi} is +.026 over \textbf{MultiDS}).  However, the small difference in performance shows that even by providing additional out-of-language training data, reasonable performance can be achieved even with reduced in-language training.

\section{Zero-shot Language Transfer}\label{sec:zero_shot}
\showboxdepth=\maxdimen
\showboxbreadth=\maxdimen

\begin{table}
\centering
\subfloat{%
\centering
\begin{tabular}{@{}cc|rrrr@{}} 
&    \multicolumn{5}{c}{Evaluation Language} \\
  & & en & zh & es & \\ \cmidrule{2-5}
\multirow{12}{*}{\rotatebox[origin=c]{90}{\parbox[c]{4cm}{\centering Training Setting}}} & Multi & 0.66 & 0.79 & 0.74 & \\ \cmidrule{2-5}
& en & .00 & \minus.03 & \minus.02 & \\
& zh & \minus.05 & .00 & \minus.03 & \\
& es & \minus.06 & \minus.06 & \minus.03 & \\ \cmidrule{2-6} \morecmidrules \cmidrule{2-6}
& & ar & fa & ko & ru \\ \cmidrule{2-6}
& Multi & 0.72 & 0.86 & 0.88 & 0.87 \\ \cmidrule{2-6}
& ar & +.03 & \minus.08 & \minus.08 & \minus.05 \\
& fa & \minus.14 & \minus.04 & \minus.16 & \minus.10 \\
& ko & \minus.20 & \minus.13 & \minus.03 & \minus.09 \\
& ru & \minus.20 & \minus.08 & \minus.13 & \minus.03 \\ 
\end{tabular} 

\label{table:zs_wiki}%
}
\caption{$\Delta$\Fone for each single-language trained model, compared to a multilingually-trained model, for each evaluation language. Each column is an evaluated language, and each row is a training setting. While models trained on the target language perform best, many monolingually-trained models perform well on unseen languages.}
\label{tab:zero_shot}
\end{table}
\begin{table}
\centering
\begin{tabular}{@{}l|rr|cc|cc@{}}
& \multicolumn{2}{c|}{ en } & \multicolumn{2}{c|}{ zh } & \multicolumn{2}{c}{ es } \\
 & avg & \Fone& avg & \Fone & avg & \Fone \\ \midrule
{name} & 0.59 & 0.70 & 0.45 & 0.71 & 0.42 & 0.73 \\ \midrule
{+cont} & +.12 & +.05 & +.22 & +.05 & +.14 & +.05 \\
{+type} & +.03 & +.01 & +.10 & \minus.02 & +.03 & \minus.03 \\
{all} & +.12 & +.05 & +.26 & +.08 & +.19 & +.06 \\ 
\end{tabular}
\caption{ English-only trained $\Delta$micro-average and $\Delta$\Fone when using a subset of linker features, compared to the name-only model for each language in the Development set. The name component of the model has the highest performance impact, but context also leads to better performance in almost all cases.}
\label{tab:zs_ablation}
\end{table}

\begin{table}
    \centering
\begin{tabular}{cc|rrrr}
BERT & Lang & micro & prec. & recall & \Fone \\ \midrule
en & en & \minus.07 & +.17 & \minus.13 & \minus.03 \\
en & es & \minus.01 & .00 & \minus.02 & \minus.01 \\
ar & ar & \minus.08 & \minus.08 & \minus.03 & \minus.06 \\
ar & fa & \minus.09 & \minus.05 & \minus.08 & \minus.06 \\
\end{tabular}
    \caption{Change in performance for monolingually-trained models using monolingually-trained BERT models, compared to monolingually-trained models using mBERT.}
    \label{tab:monobert}
\end{table}

Encouraged by the results on multilingual training, we explore performance in a zero-shot setting. 
How does a model trained on a single language perform when applied to an unseen language? We consider all pairs of languages, \ie train on each language and evaluate on all others in the same dataset\footnote{Work in Cross-language entity linking~\cite{upadhyay-etal-2018-joint, Tsai2016} has done similar evaluations, but focus on using external large data sources (Wikipedia) to train their models.}.

Table \ref{tab:zero_shot} shows the change in \Fone for monolingually-trained models compared to multilingual models. While zero-shot performance does worse than a model with access to within-language training data, the degradation is surprisingly small:  often less than 0.1 \Fone. 
For example, a model trained on all 3 \textbf{TAC} languages achieves an \Fone of 0.79 on Chinese, but if only trained on English, achieves an \Fone of 0.76.  This pattern is consistent across both models trained on related languages (Arabic $\rightarrow$ Farsi, loss of 0.08 \Fone), and on unrelated languages (Russian $\rightarrow$ Korean, loss of 0.13 \Fone).

{\bf Analysis.} Why does zero-shot language transfer do so well for cross-language entity linking? What challenges remain to eliminate the degradation in performance from zero-shot transfer?

We answer these questions by exploring the importance of each component of our cross-language ranking system: mention string, context, and type. We conduct ablation experiments investigating the performance loss from removing these information sources. We then evaluate each model in an English-trained zero-shot setting.
First, we train a zero shot model using only the mention text and entity name. We then compare the performance change that results from adding the context, the type, and both context and type (all features). 

Table \ref{tab:zs_ablation} shows that comparing the name and mention text alone accounts for most of the model's performance, a sensible result given that most of the task involves matching entity names. We find that context accounts for most of the remaining performance, with type information having a marginal effect.
This highlights the importance of the multilingual encoder, since both name and context rely on effective multilingual representations.

Separately, how does using a multilingual transformer model, such as mBERT, affect the performance of our ranker?  First, it is possible that using a monolingual linker with a BERT model trained only on the target language would improve performance, since such a model does not need to represent several languages as the same time.  As shown in Table \ref{tab:monobert}, model performance for these settings is largely worse for English-only and Arabic-only \cite{safaya-etal-2020-kuisail} models when compared to using mBERT, with the exception that precision increases significantly for English.  Second, perhaps a monolingual linker with a BERT model trained only on a related language -- \eg English BERT for Spanish, Arabic BERT for Farsi -- would produce acceptable results.  Again, as shown in Table \ref{tab:monobert}, the performance is most often worse, illustrating that mBERT is an important aspect of the linker's performance.

\begin{table*}
\begin{tabular}{@{}rr|rr|rrc|crr|rr|rr@{}}
\multicolumn{2}{c|}{ en } & \multicolumn{2}{c|}{ zh } & \multicolumn{2}{c}{ es } & && \multicolumn{2}{c|}{ en } & \multicolumn{2}{c|}{ zh } & \multicolumn{2}{c}{ es } \\
avg & \Fone & avg & \Fone & avg & \Fone & & & avg & \Fone & avg & \Fone & avg & \Fone \\ \cmidrule{1-6} \cmidrule{9-14}
0.64 & 0.75 & 0.51 & 0.69 & 0.53 & 0.75 & Rand & Tail & 0.53 & 0.66 & 0.45 & 0.66 & 0.42 & 0.70 \\ \cmidrule{1-6} \cmidrule{9-14}
.00 & \minus.01 & +.07 & +.02 & \minus.02 & \minus.02 & \multicolumn{2}{c}{ w/ Name } & \minus.02 & \minus.02 & +.02 & +.01 & \minus.01 & \minus.01 \\
.00 & +.01 & +.06 & +.04 & +.01 & +.02 & \multicolumn{2}{c}{ w/ Pop-Train} & \minus.02 & +.04 & .00 & +.07 & \minus.01 & +.06 \\
+.04 & +.03 & +.12 & +.06 & +.10 & +.06 & \multicolumn{2}{c}{w/ Pop-All} & +.13 & +.10 & +.20 & +.11 & +.22 & +.10 \\
\end{tabular} 
\caption{For each proposed Name matching or popularity re-ranking model, the change in performance ($\Delta$\Fone and $\Delta$micro-average) compared to the original \textbf{Rand} (left) and \textbf{Tail} (right) models.  While the name matching increased performance somewhat, the additional of popularity was more impactful.}
\label{tab:zs_improv}

\end{table*}

\begin{table}
\centering
\begin{tabular}{@{}l|cc|cc|cc@{}}
& \multicolumn{2}{c|}{ en } & \multicolumn{2}{c|}{ zh } & \multicolumn{2}{c}{ es } \\
& avg & \Fone & avg & \Fone & avg & \Fone \\ \midrule
Multi & 0.70 & 0.73 & 0.77 & 0.81 & 0.68 & 0.82 \\ \midrule
Rand & \minus.04 & -.02 & \minus.26 & \minus.12 & \minus.15 & \minus.07 \\
N-1 & +.01 & +.02 & \minus.04 & \minus.02 & \minus.08 & \minus.03 \\
N-1U & \minus.24 & -.14 & \minus.49 & \minus.22 & \minus.38 & \minus.19 \\
Tail & \minus.16 & -.08 & \minus.31 & \minus.15 & \minus.26 & \minus.12 \\ 

\end{tabular} 
\caption{For each of the English-only training data subsets described in \S \ref{sec:ent_pop}, $\Delta$Micro-average and $\Delta$\Fone compared to the full \textbf{Multi} model. Models that see even a single example of an entity (\eg \textbf{N-1}) outperform models that see a portion (\eg \textbf{Tail}) or none (\eg \textbf{N-1U}).}
\label{tab:ent_exp}
\end{table}

\section{Improving Zero-shot Transfer}
\subsection{Name Matching Objective}\label{sec:aux}
\begin{CJK*}{UTF8}{gbsn}

Given the importance of matching the mention string with the entity name, will improving this component enhance zero-shot transfer? While obtaining within-language entity linking data isn't possible in a zero-shot setting, we can use pairs of translated names, which are often more easily available \cite{irvine2010transliterating,peng-etal-2015-empirical}. 
Since Chinese performance suffers the most zero-shot performance reduction when compared to the multilingual setting, we use Chinese English name pair data~\cite{huang_2005} to support an auxiliary training objective. An example name pair: ``巴尔的摩－俄亥俄铁路公司'' and \textit{Baltimore \& Ohio Railroad}.
\end{CJK*} 

We augment model training as follows.
For each update in a mini-batch, we first calculate the loss of the subset of the model that scores the mention string and entity name on a randomly selected pair $k=25,000$ of the Chinese/English name pair corpus.
We score the Chinese name $z$ and the correctly matched English name $e_+$ pair, and separately score the same Chinese name paired with $n$ negatively sampled English names $e_{-}$.  We create representations for both $z$ and $e$ using the method described for names in \S \ref{sec:rep} which are passed to the name-only hidden layer.  We add a matching-specific hidden layer, which produces a score. We apply the hinge loss between positive and negative examples,
\begin{multline*}
    \mathit{N}(\theta) = \mathbf{max}\{0,\epsilon-(S(\{z,e_{+}\};\theta)-\\ \mathbf{max}\{S(\{z,e_{0-}\};\theta) \dots S(\{z, e_{n-}\};\theta)\}\}
\end{multline*}
The name pair loss is then multiplied by a scalar $\lambda = 0.5$ and added to the loss described in \S \ref{sec:training}.
The resulting loss $\mathit{L_{joint}}(\theta) = (\lambda * \mathit{N}(\theta)) + \mathit{L}(\theta)$ is jointly minimized.
After training, we discard the layer used to produce a score for name matches. This procedure still only uses source language entity linking training data, but makes use of auxiliary resources to improve the name matching component, the most important aspect of the model.

We analyze the resulting performance by considering modifications to our English-only training setting, which are designed to replicate scenarios where there is little training data available.
To show the effect of a smaller training corpus, we select a random 50\% of mentions, partitioned by document (\textbf{Rand}). To show the importance of training on frequently occurring entities, we select 50\% of mentions that are linked to the least frequent entities in the English dataset (\textbf{Tail}).  

Table \ref{tab:zs_improv} shows the results on each of the three development \textbf{TAC} languages compared to the \textbf{Multi} model. 
For the \textbf{Rand} training set, we see a large improvement in Chinese micro-average and a small one in \Fone, but otherwise see small reductions in performance. In the \textbf{Tail} training setting, a similar pattern occurs, with the exception that Chinese is less improved than in \textbf{Rand}.  Overall, performance loss remains from zero-shot transfer which suggests that improvements need to be explored beyond just name matching.


\subsection{Entities}\label{sec:ent_pop}
Another possible source of zero-shot degradation is the lack of information on specific entities mentioned in the target language. For entity linking, knowledge of the distribution over the ontology can be very helpful in making linking decisions.  While zero-shot models have access to general domain text, \ie news, they often lack text discussing the same entities. For example, some entities that only occur in Chinese (231 unique entities in \textbf{Dev}), such as the frequently occurring entity \textit{Hong Kong}, have a number of similar entities and thus are more challenging to disambiguate.

We measure this effect through several diagnostic experiments where we evaluate on the development set for all languages, but train on a reduced amount of English training data in the following ways:
In addition to the \textbf{Rand} and \textbf{Tail} settings, we sample a single example mention for each entity (\textbf{N-1}), resulting in a much smaller training as compared to those datasets. We also take \textbf{N-1} and remove all evaluation set entities (\textbf{N-1U}), leaving all evaluation entities unseen at train time.

Table \ref{tab:ent_exp} reports results on these reduced training sets. All languages use a $-1$ NIL threshold. Compared to the multilingual baseline (\textbf{Multi}) trained on all languages, there is a decrease in performance in all settings. Several patterns emerge.
First, the models trained on a subset of the English training data containing more example entities - \eg \textbf{N-1} - have much higher performance than the models that do not.  This is true even in non-English languages. Unobserved entities do poorly at test time, suggesting that observing entities in the training data is important.

However, a mention training example can improve the performance of a mention in another language if linked to the same entity, which suggests that this provides the model with data-specific entity information. Therefore, the remaining zero-shot performance degradation can be largely attributed not to a change in language, but to a change in topic, \ie what entities are commonly linked to in the data.
This may also explain why although the name matching component is so important in zero-shot transfer, our auxiliary training objective was unable to fully mitigate the problem. The model may be overfitting to observed entities, forcing the name component to memorize specific names of popular entities seen in the training data.
This means we are faced with a topic adaptation rather than a language adaptation problem.

\label{sec:pop}

We validate this hypothesis by experimenting with information about entity popularity. 
Will including information about which entities are popular improve zero-shot transfer?
We answer this question by re-ranking the entity linker's top ten predicted entities using popularity information, selecting the most most popular entity from the list. Adding this feature into the model and re-training did not lead to a significant performance gain.
We define the popularity of an entity to be the number of times it occurred in the training data.  We report results for two popularity measures--one using the popularity of the English subset of the data used for training, and one using all of the training data (including for Spanish and Chinese).  

Table \ref{tab:zs_improv} shows that both strategies improve \Fone, meaning that a missing component of zero-shot transfer is information about which entities are favored in a specific dataset. The gain from using popularity estimated from the training data only is smaller than using the popularity data drawn from all of \textbf{TAC}. With more accurate popularity information, we can better mitigate loss.

Several patterns emerge from most common corrections made with the Population reranking for \textbf{Tail}, included in Table \ref{tab:app_reranking}. Many errors arise from selecting related entities that are closely related to the correct entity -- for example, \textit{United States Congress} instead of the \textit{United States of America}.  Additionally, people with similar names are often confused (e.g. \textit{Edmund Hillary} instead of \textit{Hillary Clinton}).  Finally, many appear to be annotation decisions -- often both the original prediction (e.g. \textit{Islamic State}) and the corrected popular prediction (e.g. \textit{Islamic State of Iraq and Syria}) appear reasonable choices.  While most corrections were in Chinese (632), some occurred in both English (419) and Spanish (187).  These errors -- especially those in English -- illustrate that much of the remaining error is in failing to adapt to unseen entities.

\section{Conclusion}
We demonstrate that a basic neural ranking architecture for cross-language entity linking can leverage the power of multilingual transformer representations to perform well on cross-lingual entity linking. Further, this enables a multilingual entity linker to achieve good performance, eliminating the need for language-specific models.
Additionally, we find that this model does surprisingly well at zero-shot language transfer. We find that the zero-shot transfer loss can be partly mitigated by an auxiliary training objective to improve the name matching components. However, we find that the remaining error is {\em not} due to language transfer, but to topic transfer. Future work that improves zero-shot transfer should focus on better ways to adapt to entity popularity in target datasets, instead of relying on further improvements in multilingual representations.  Focusing on adapting to the topic and entities present in a given document is critical.  This could be accomplished by adding a document-level representation or by leveraging other mentions in the document. English-focused work on rare entity performance \cite{orr2020bootleg,jin2014entity} may provide additional direction.

\bibliographystyle{acl_natbib}
\bibliography{anthology,emnlp2020}




\begin{appendix}
\section{Architecture information}\label{app:arch}

\begin{table}[h]
\centering
\begin{tabular}{l|p{3.5cm}}
Parameter & Values \\ 
Context Layer(s) & [768], \textbf{[512]}, [256], [512,256] \\
Mention Layer(s) & [768], \textbf{[512]}, [256], [512,256] \\
Type Layer & [128], \textbf{[64]}, [32], [16] \\
Final Layer(s) & \textbf{[512,256]}, [256,128], [128,64], [1024,512], [512], [256] \\
Dropout probability & 0.1, \textbf{0.2}, 0.5 \\
Learning rate & 1e-5, 5e-4, \textbf{1e-4}, 5e-3, 1e-3 \\
\end{tabular} 
\renewcommand\thetable{6} 
\caption{To select parameters for the ranker, we tried 10 random combinations of the above parameters, and selected the configuration that performed best on the TAC development set.  The selected parameter is in bold. We report results after training for 500 epochs for TAC and 800 for Wiki.  The full TAC multilingual model takes approximately 1 day to train on a single NVIDIA GeForce Titan RTX GPU, including candidate generation, representation caching, and prediction on the full evaluation dataset.}
\end{table}

\section{Dataset Details}\label{app:triage}
The NIL threshold is selected based on the development\textbf{TAC} dataset. Unless noted, we use $-0.8$ for English and $-1$ otherwise.

\textbf{TAC:} The training set consists of 30,834 mentions (6,857 NIL) across 447 documents. We reserved a randomly selected 20\% of these documents as our development set, and will release development splits.
The evaluation set consists of 32,459 mentions (8,756 NIL) across 502 documents. A mention is linked to NIL if there is no relevant entity in the KB, and the KB is derived from a version of BaseKB.  

\textbf{TAC Triage:} We use the system discussed in  for both the \textbf{TAC} and \textbf{Wiki} datasets.  However, while the triage system provides candidates in the same KB as the \textbf{Wiki} data, not all entities in the \textbf{TAC} KB have Wikipedia page titles.  Therefore, the \textbf{TAC} triage step requires an intermediate step - using the Wikipedia titles generated by triage ($k=10$), we query a Lucene database of BaseKB for relevant entities.  For each title, we query BaseKB proportional to the prior provided by the triage system, meaning that we retrieve more BaseKB entities for titles that have a higher triage score, resulting in $l=200$ entities.  First, entities with Wikipedia titles are queried, followed by the entity name itself.  If none are found, we query the mention string - this provides a small increase in triage recall.  This necessary intermediate step results in a lower recall rate for the \textbf{TAC} dataset (85.1\% for the evaluation set) than the \textbf{Wiki} dataset, which was 96.3\% for the evaluation set .

\textbf{Wiki:} Some BaseKB entities used in the \textbf{TAC} dataset have Wikipedia links provided; we used those links as seed entities for retrieving mentions, retrieving mentions in proportion to their presence in the \textbf{TAC} dataset, and to sample a roughly equivalent number of non-TAC entities. We mark 20\% of the remaining mentions as NIL. In total, we train and evaluate on 5,923 and 1,859 Arabic, 3,927 and 1,033 Farsi, 5,978 and 1,694 Korean, and 5,337 and 1,337 Russian mentions, respectively.

\begin{table*}[t]

    \centering
\begin{tabularx}{\linewidth}{ X |l| r }
Original Prediction & Popular Correction & Count \\ \midrule
United States   Department of State & United States of America & 146 \\
united\_states\_congress & United States of America & 121 \\
Soviet Union & Russian & 57 \\
Central Intelligence Agency & United States of America & 41 \\
healthcare\_of\_cuba & Cuba & 36 \\
islamic\_state & Islamic State of Iraq and Syria & 33 \\
edmund\_hillary & First lady Hillary Rodham Clinton & 32 \\
United States Department of Defense & United States of America & 32 \\
Tamerlan Tsarnaev & Dzhokhar A. Tsarnaev & 27 \\
Carl Pistorius & Oscar Leonard Carl Pistorius & 23 \\
CUBA\_Defending\_Socialism\_ ... documentary & Cuba & 22 \\
Barack Obama Sr. & Barack Hussein Obama II & 18 \\
Iraq War & Iraq & 14 \\
Dzhokhar Dudayev & Dzhokhar A. Tsarnaev & 13 \\
Sumter County / Cuba town & Cuba & 13 \\
United States Army & United States of America & 13 \\
military\_of\_the\_united\_states & United States of America & 13 \\
Republic of Somaliland & Somalian & 13 \\
ISIS & Islamic State of Iraq and Syria & 13 \\
Islamic\_State\_of\_Iraq\_and\_Syria & Islamic State of Iraq and Syria & 12 \\
National Assembly of People's Power & Cuba & 11 \\
Sara Netanyahu & Benjamin Netanyahu & 10
\end{tabularx}
    \caption{All pairs of original prediction and popular prediction altered by the reranking procedure described in Section 7.2, for the \textbf{Tail} model}
    \label{tab:app_reranking}
\end{table*}
\end{appendix}

\end{document}